\documentclass[10pt,twocolumn,letterpaper]{article}

\usepackage{wacv}
\usepackage{times}
\usepackage{epsfig}
\usepackage{graphicx}
\usepackage{amsmath}
\usepackage{amssymb}

\usepackage{multirow}
\usepackage{booktabs}

\def\wacvPaperID{0658} 

\wacvfinalcopy 

\ifwacvfinal
\def\assignedStartPage{9876} 
\fi

\ifwacvfinal
\usepackage[breaklinks=true,bookmarks=false]{hyperref}
\else
\usepackage[pagebackref=true,breaklinks=true,colorlinks,bookmarks=false]{hyperref}
\fi

\ifwacvfinal
\else
\fi

\begin{document}

\title{ Do not Forget to Attend to Uncertainty while Mitigating Catastrophic Forgetting}

\author{Vinod K Kurmi\\
IIT Kanpur\\
{\tt\small vinodkk@iitk.ac.in}
\and
Badri N Patro\thanks{Currently working at Google}
\\
IIT Kanpur\\
{\tt\small badri@iitk.ac.in}
\and
Venkatesh K Subramanian\\
IIT Kanpur\\
{\tt\small venkats@iitk.ac.in}
\and
Vinay P Namboodiri\\
University of Bath\\
{\tt\small vpn22@bath.ac.uk}
}

\maketitle

\begin{abstract}
One of the major limitations of deep learning models is that they face catastrophic forgetting in an incremental learning scenario.  There have been several approaches proposed to tackle the problem of incremental learning. Most of these methods are based on knowledge distillation and do not adequately utilize the information provided by older task models, such as uncertainty estimation in predictions. The predictive uncertainty provides the distributional information can be applied to mitigate catastrophic forgetting in a deep learning framework.  In the proposed work, we consider a Bayesian formulation to obtain the data and model uncertainties. We also incorporate self-attention framework to address the incremental learning problem. We define distillation losses in terms of aleatoric uncertainty and self-attention. In the proposed work, we investigate different ablation analyses on these losses. Furthermore, we are able to obtain better results in terms of accuracy on standard benchmarks.

\end{abstract}

\section{Introduction}
Deep neural networks have shown impressive results on a variety of computer vision tasks. However, these models have been observed to be not well suited for generalization in terms of tasks. This has been particularly observed, for instance, in seminal work by Kirkpatrick {\it et al.} \cite{kirkpatrick2017overcoming} that the deep learning models can face catastrophic forgetting. Learning a model on a task `A' results in an optimal model being obtained for task `A'. However, using this same model to solve for task `B' results in the model converging to new optima that are significantly worse for solving task `A'. This is in contrast to humans, where we are able to learn new tasks without forgetting. In order to obtain deep learning networks that can continually learn without forgetting, we need to solve this problem. There have been a number of methods proposed towards solving this problem, such as those based on regularization~\cite{kirkpatrick2017overcoming, zenke2017continual} or those based on memory replay~\cite{wu2018memory, aljundi2018memory}. However, the methods proposed are not focused on the uncertainty associated with the distillation. One approach that allows us to understand the working of the method is by considering the uncertainty distribution. Deep learning models usually consider the point estimate for predicting instead of adopting a probabilistic distribution-based approach. In the probabilistic approach, a fully Bayesian treatment is intractable in general. However, work by Gal and Ghahramani~\cite{gal2016dropout} and other related subsequent works~\cite{kendall2017uncertainties,kendall2018multi} suggest that it is possible to obtain the uncertainty of a model in its parameters or the data distribution during prediction. We adopt this as it is particularly relevant in the case of incremental learning. By incorporating uncertainty, we can ensure that the model will initially be uncertain about the new task, and as it is trained, its uncertainty reduces. Moreover, we consider the uncertainty based distillation loss between tasks that ensure the method distills the uncertainty along with the prediction.


Another approach that allows for interpretability is considering the attention and visual explanation based models. Attention-based methods have been very popular in machine translation~\cite{vaswani_nips2017}, visual question answer~\cite{yang2016stacked}, and generative adversarial networks (GANs)~\cite{zhang2019self}. In our work, we consider self-attention based methods~\cite{zhang2019self} for incremental learning. We particularly consider the distillation loss between tasks that ensure the method attends to similar discriminative regions. Moreover, obtaining the attention-based distillation can result in improved incremental learning techniques being developed.

Our work is based on a baseline that is a variant of learning without forgetting (LwF)~\cite{li2017learning}. Particularly we propose a variant that incorporates a distillation loss between the previous task and the new task. However, all the samples used for training the previous task would not be available while training for a new task. Therefore, we consider a method that retains a small amount of data from the previous task and uses distillation loss to retain the knowledge while learning a new task~\cite{rebuffi2017icarl}.

 In the data-based incremental learning setting, previous works have proposed different distillation based models to
preserve the old task information while training to solve for the current task. The main problem with these methods is that they distill the knowledge of previous data from the previous model's softmax output for a given input of current task data. Since the predictions of deep learning models are overconfident for any prediction, it means that any single class has probability near one, while other classes have almost zero probability: even when the prediction is wrong~\cite{malinin2018predictive}.
 To solve this problem, it has been suggested that high temperatures can be used~\cite{hinton2015distilling} for calculating the distillation. However, this does not truly reflect the uncertainty in prediction.  In incremental learning, the model distills knowledge from the next task, for which it is not trained. Thus it should not be overconfident for one class; instead, the prediction should be distributed among all the classes. There will be uncertainty present in the prediction. It means this distillation can not truly reflect the older class information properly. So the  softmax based distillation loss is not sufficient to preserve the previous task knowledge in the current model. Thus, in the proposed method, we use uncertainty-based distillation to reflect information about the older task data.

Another problem with incremental learning is that not all the information provided by the previous model is useful for preserving the older task data. The attention-based method can further improve the performance of incremental learning. In~\cite{dhar2019learning} authors propose a GradCAM~\cite{selvaraju2017grad} based attention distillation for class incremental learning. GradCAM based attention also depends upon the predicted class label of the previous model.
In contrast to the predicted label based attention, we propose self-attention based distillation for the incremental learning. The idea of the self-attention module is to adapt the non-local model information using the residual connections.

In the next section, we discuss the related work. The proposed method is presented in detail in section 3. In section 4, we present thorough empirical analysis and results for the various evaluations on standard CIFAR-100 and ImageNet datasets. We finally conclude our paper with directions for future work. 
The major contributions of the proposed work are as follows:
\begin{itemize}
    \item We propose an uncertainty and attention based distillation in incremental learning problem.
    \item We show that by transferring the uncertainty, student model captures  more information about the teacher model.
    \item Attention framework further reduce the catastrophic forgetting.
    \item We thoroughly analyze the proposed approach through ablation analysis and visualization.
\end{itemize}


\section{Related Work}


Many methods have been proposed to solve catastrophic forgetting in the field of incremental or lifelong learning. One of the basic and intuitive methods among them is LwF~\cite{li2017learning}. For each task, this method uses the same set of feature extractors and task-specific layers for classification. In general, the incremental learning methods can be broadly divided into three categories: \textbf{i)} task-specific methods \textbf{ii)} regularization based methods and \textbf{iii)} methods based on memory replay. The task-specific methods deal with adding a separate model or layer for each new task~\cite{masse2018alleviating,li2017learning}. Regularization based methods use a regularization term to avoid changing those parameters that are sensitive to old tasks. Elastic weight consolidation (EWC)~\cite{kirkpatrick2017overcoming} is a regularization based model that quantifies the importance of weights for previous tasks, and selectively alters the learning rates of weights. Following EWC, the synaptic consolidation strength is measured by~\cite{zenke2017continual} in an online fashion and used it as regularization. In Incremental Moment Matching (IMM)~\cite{lee2017overcoming}, a trained network on each
task is preserved, and the networks are later merged into one at the end of all the tasks. The above approaches focus on adjusting the network weights. In~\cite{rajasegaran2019random}, the authors use the dropout based model selection for incremental learning.

The third category is data-based methods that use the manifold distance between the model trained on old tasks and model trained on the current task. This distance is minimized by suitable loss functions. The most commonly used manifold and loss is the distillation loss~\cite{hinton2015distilling}.  These methods need feature for obtaining the loss function; thus, a small amount of data of previous task is kept.  Therefore, the amount of knowledge kept by knowledge distillation depends on the degree of similarity between the data distribution used to learn the previous tasks in the previous stages.  Other data-generation based methods~\cite{lesort2019generative,rannen2017encoder,shin2017continual,van2018generative,wu2018incremental} use generative adversarial network~\cite{goodfellow2014generative} and by replying the total data, new model is trained. Gradient Episodic Memory (GEM)~\cite{lopez2017gradient,rebuffi2017icarl} stores a subset of training data. The End-to-End incremental learning (E2E) framework~\cite{castro2018end} uses a task-specific distillation loss. In~\cite{hou2018lifelong}, a distillation and retrospection (D+R) used to preserve the previous task information. Recently in~\cite{lee2019overcoming} the external data along with global distillation (GD) loss is used to improve the incremental learning performance.  Another method iCaRL~\cite{rebuffi2017icarl} also considers the distillation loss for incremental learning. Other data-based incremental methods are proposed in~\cite{li2018supportnet}. 
 Similarly,~\cite{schwarz2018progress} uses the two teacher models for model-based incremental learning. In~\cite{hou2018lifelong} authors also use the distillation loss in task incremental learning setting.
 These methods consider the distillation loss without incorporating the uncertainty value.
 
 \begin{figure*}
 \centering
    \includegraphics[width=0.7975\textwidth]{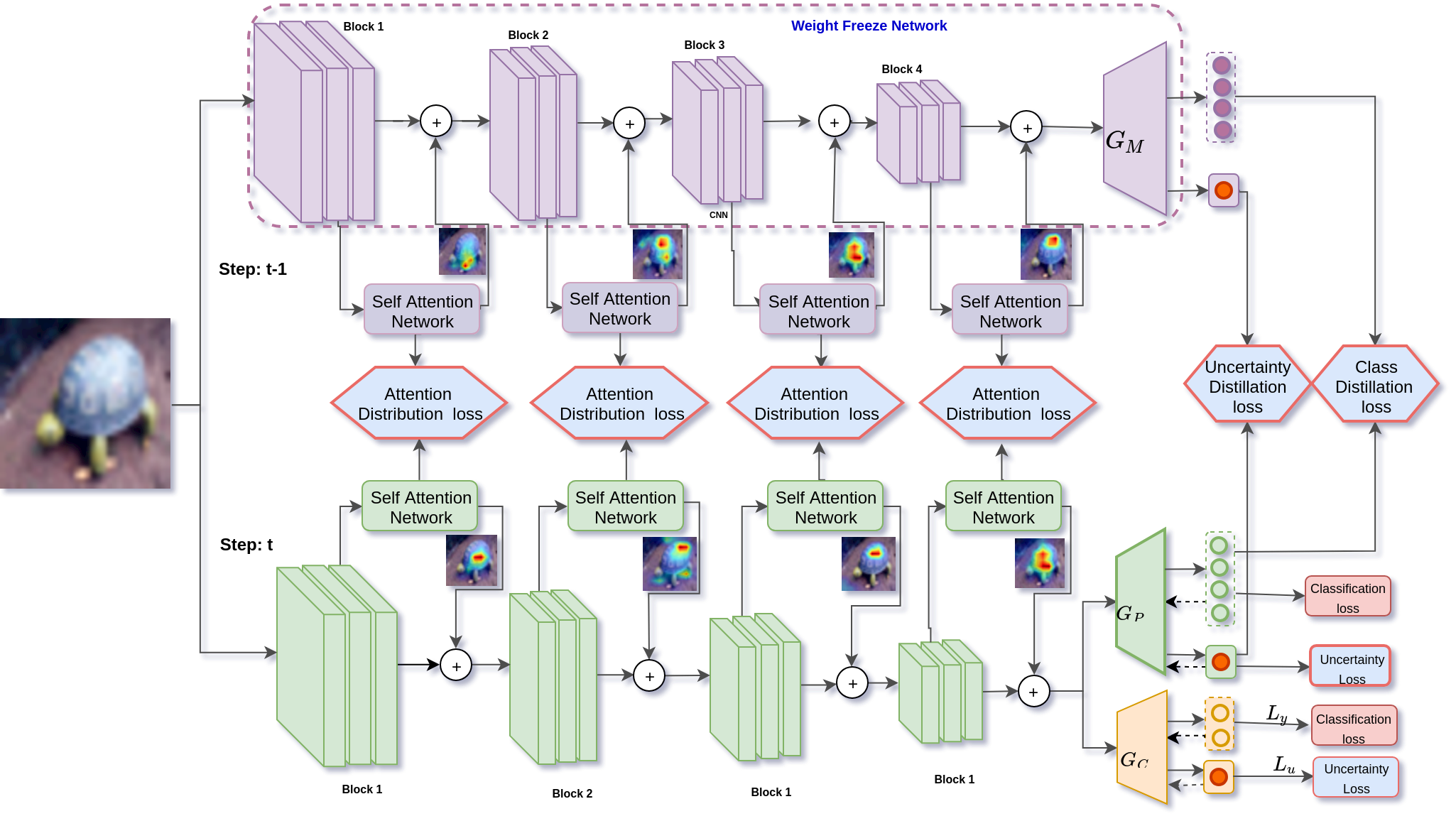}
      \caption{Overview of the uncertainty distillation and self-attention in incremental learning. The upper part of the figure represents the model that is trained on the previous tasks and is set to be frozen. $G_M$ is pretrained classifier trained on the previous tasks, $G_P$, and $G_c$ are the classifiers of the current model for training the old tasks classes and new task classes, respectively. }
      \label{fig:main_diagram}
 \end{figure*}

 Recently some attention based methods have been used in incremental learning~\cite{dhar2019learning,serra2018overcoming}. A GradCAM based attention used in~\cite{dhar2019learning} , while authors in~\cite{serra2018overcoming} use a attention module to avoid  catastrophic forgetting. In~\cite{kochurov2018bayesian}, a Bayesian framework is discussed for the incremental learning problem. But these methods do not consider uncertainty for preserving the knowledge of previous tasks. Moreover, we differ by considering self-attention based methods for obtaining the attention \cite{vaswani_nips2017,zhang2019self}. This allows us to learn the attention regions that are relevant for the task, unlike the work that uses GradCAM~\cite{selvaraju2017grad} that requires the true class label for obtaining a visualization. 
 
 Recently, work by~\cite{zhao2020maintaining} utilizes knowledge distillation to maintain discrimination within old classes to avoid catastrophic forgetting. The adversarial continual learning~\cite{ebrahimi2020adversarial} method learns a disjoint representation for task-invariant and task-specific features. PODNet~\cite{douillardpodnet} approaches incremental learning as representation learning, with a distillation loss that constrains the evolution of the representation. iTaML~\cite{rajasegaran2020itaml} proposes a task agnostic meta-learning approach that seeks to maintain an equilibrium between all the encountered tasks. ScaIL~\cite{belouadah2020scail} discusses an efficient scaling of past classifiers’ weights to make them more comparable to those of new classes in incremental learning scenarios.

\section{Method}
We propose a class incremental learning framework based on the uncertainty distillation and multi-stage self-attention, where class data come in sequentially in a batch of classes.  The model is illustrated in Fig.~\ref{fig:main_diagram}.  

\subsection{Problem Formulation}

Let us assume  dataset $D$ consists of pairs of images $x$ and its label $y$, i.e $D \in (x,y)$.
In the proposed framework, we assume there are a total of $T$ tasks i.e. the classification dataset is divided into $T$ parts $D_1$, $D_2$, ...,$D_T$. where each $D_t=\{x_j^t,y_j^t\}_{j=1}^{n_t}$. Our goal is to learn a classifier $C_T$  that can correctly classify all the images of all the tasks.
Let's assume $D^{old}_t$ contain a fixed number of representative samples from the old dataset, i.e. $D^{old}_t \subset D_1$ $\cup$ $D_2$ $\cup$ ...$\cup$ $D_{t-1}$. 
In the task increment learning problem, at timestamp $t$, we can  access the dataset of the current task and the $({t-1})^{th}$ representative data, i.e. $D^{train}_t=D_t \cup D^{old}_t$. 

The classifiers of the model can be divided into three types: \textbf{i)} previous step's classifier ($G_m$), parameters $\theta_m$, trained on the previous task(s) data, \textbf{ii)} the current model classifier ($G_p)$, parameters $\theta_p$, trained on the current task data of previous classes, and \textbf{iii)} current model classifier ($G_c)$, parameters $\theta_c$, trained on new classes at $t^{th}$ stage. At next $({t+1})^{th}$ stage the models parameters $\theta_m$  will be $\theta_p \cup \theta_c$. There are three-stage attention networks in the feature extractor, which has the parameters $\theta_f$.
 The model is trained using the knowledge distillation from the previous model as well as the cross-entropy loss for the current task data.
\subsection{Knowledge Distillation}
Knowledge distillation~~\cite{hinton2015distilling} is obtained by minimizing the loss function between the distribution of class probabilities predicted by the teacher and student model. In the deep learning models, the correct class's predicted probability is very high, while other class probabilities are nearly equal to zero. If the ground truth label is provided to the student network, then the teacher model also predicts the same output. Thus the teacher model does not provide any useful information to students. This problem has been tackled using a softmax temperature by Hinton {\it et al.} ~\cite{hinton2015distilling}. 
     
    \begin{equation}
     p_{c,n}=\frac{\exp \left({q_{c,n}} / \tau\right)} {\sum_{j=1}^{C} \exp \left(q_{j,n} / \tau\right)}
     \end{equation}

 $q_{c,n}$ is the logit value for $c^{th} \in \mathcal{C}$ class for input $x_n \in \mathcal{D}$ from the model $M_t$. $\tau$ is the temperature value.   
 Then distillation loss is defined as follows: 
 \begin{equation}
    \mathcal{L}_{D}(\theta_s,\mathcal{D}, \mathcal{C})= \frac{1}{|\mathcal{D}|}{\sum_{n=1}^{N}}\sum_{c=1}^{C} -p_{c,n} \cdot \log \left(\hat{y}_{c,n}\right)
 \end{equation}
 
 $\hat{y}_{c,n} $ is the predicted $c^{th}$ class logit for student network($M_s(:, \theta_s)$) for input $x_n$. $N$ is the total number of samples and $C$ is the total number of classes.

    
\subsection{Distillation in Incremental Learning}    

In the incremental learning setting, suppose the model $M_{prev}$ is trained on the total $|\mathcal{C}_{prev}|$ classes and at $t^{th}$ task, there are new $|\mathcal{C}_{t}|$ classes introduced. The objective is to train the current model $M_t$  for all classes i.e $\mathcal{C}_{cur}= \mathcal{C}_{t} \cup  \mathcal{C}_{prev}$. The current model $M_t$ is defined by the feature extractor ($M^f_t, \theta_f$), old class classifier ($G_p,\theta_p$)  and new class classifier ($G_c,\theta_c$).

For the old task model, distillation loss at $t^{th}$ stage is given as: 

\begin{equation}
    \mathcal{L}_{\mathrm{D}}(\theta_f,\theta_p ; \mathcal{C}_{prev})  = \frac{1}{|{D^{train}_t}|} \sum_{x \in {D}^{train}_t} \sum_{y \in \mathcal{C}_{prev}}\left[- p_{c} \log p(y | x )\right] 
     \label{eq:dis}
\end{equation}

$p_c$ is the softmax output of the previous model, and $p(y|x)$ is the output of the model $G_p$. Since the previous task model trained on the total $|\mathcal{C}_{prev}|$, so distillation is calculated for the old task class logit value. For the current task data, the current model $M_t$ is trained using the cross-entropy loss as given by the equation;

\begin{equation}
   \mathcal{L}_{\mathrm{C}}(\theta_f,\theta_c; \mathcal{C}_{t})  = \frac{1}{|{D^{train}_t}|} \sum_{x \in {D}^{train}_t} \sum_{y_c \in \mathcal{C}_{t}}\left[- y_c \log (\hat{y}_c )\right] 
    \label{eq:crs}
\end{equation}
where $y_c$ is the ground truth label  and $\hat{y}_c$ is the prediction of $G_c$ for the input $x$.

\subsection{Limitation of Distillation loss}
It has been shown that the deep learning models are overconfident for the wrong prediction~\cite{malinin2018predictive}; thus, the prediction probabilities do not truly reflect the knowledge. In LwM~\cite{dhar2019learning}, a GradCAM~\cite{selvaraju2017grad} based loss (a visual explanation method) is used for the distillation; it considers the class activation map for distillation. In contrast to the class activation map, the self-attention models learn the attention using the network itself, and an uncertainty estimation predicts the uncertainty associated with it, used to train the model. So the\textit{ overconfident} class activation map for given input does not affect the distillation loss.

\subsection{Uncertainty based Distillation}
In the deep learning models, the class prediction based model is highly probable for one class, and so they cannot transfer the full distributional information to the student model. In the Bayesian neural networks, they can predict the uncertainty value along with their prediction, which can be applicable for transferring more detailed knowledge. Thus, in an incremental learning setting, uncertainty distillation combined with predictions can help the model to preserve useful information from the previous tasks.

\subsubsection{Bayesian Model and Uncertainty Estimation}
Bayesian modeling is beneficial for predicting uncertainty. In the deep learning models, obtaining the probabilistic inference is an intractable problem. However, ~\cite{gal2016dropout,kendall2017uncertainties} proposed a dropout based method to approximate the posterior. We follow a similar method to obtain the Bayesian framework for our incremental learning model. In the proposed method, we work with two types of uncertainties, data uncertainty, also known as aleatoric uncertainty and model uncertainty or epistemic uncertainty.  We obtain these uncertainties in a manner similar to~\cite{kendall2017uncertainties,kyle,Patro_2019_ICCV,Kurmi_2019_CVPR}. We train the classifier to predicts output class probabilities along with aleatoric uncertainty (data uncertainty). The predictive uncertainty includes epistemic uncertainty and data uncertainty. The epistemic uncertainty results from uncertainty in model parameters. It is obtained by sampling the model using MC-dropout~\cite{gal2016dropout,kendall2017uncertainties}. The estimation of the aleatoric uncertainty in the Bayesian neural network also makes the model more robust. Thus it is better for the incremental learning setting. For obtaining the variance value, we define a variance network $G_{cv}$ for the current task model and $G_{pv}$ for the older task model. The variance network takes the feature for input to predict the variance value. We divide the model $M_t$ into feature extractor $G_f$ and classifier networks $G_c$  and $G_p$. Suppose $x_i$ is the $i^{th}$input images at current task $t$, the predicted class logits  and variance value are obtain as follows:



\noindent\begin{minipage}{.5\linewidth}
\begin{equation*}
 \hat{y}_{i}^{c}= G_c(G_f(x^t_i,\theta_f),\theta_c)
\end{equation*}
\end{minipage}%
\begin{minipage}{.5\linewidth}
\begin{equation*}
  \hat{y}_{i}^{p}= G_p(G_f(x^t_i,\theta_f),\theta_p)  \\
\end{equation*}
\end{minipage}

\begin{equation}
     (\sigma_{i}^{c})^2= G_{cv}(G_f(x^t_i,\theta_f),\theta_{cv}) 
      \label{eq:var_c}
\end{equation}
\begin{equation}
     ( \sigma_{i}^{p})^2= G_{pv}(G_f(x^t_i,\theta_f),\theta_{pv}) \\
  \label{eq:var_p}
\end{equation}



The aleatoric loss is obtained as: 
\begin{equation}
    \hat{y}_{i, s}^{c}=\hat{y}_{i}^{c}+\sigma_{i}^{c} * \epsilon^c_{s}, \quad \epsilon^c_{s} \sim \mathcal{N}(0, I)
\end{equation}
\begin{equation}
    \hat{y}_{i, s}^{p}=\hat{y}_{i}^{p}+\sigma_{i}^{p} * \epsilon^p_{s}, \quad \epsilon^p_{s} \sim \mathcal{N}(0, I)
\end{equation}
 
 \begin{equation}
     \mathcal{L}_{c a}=-\frac{1}{|{D^{train}_t}|} \sum_{x_{i} \in {D^{train}_t}} \log \frac{1}{\mathrm{T}_{\mathrm{a}}} \sum_{s=1}^{\mathrm{T}_{\mathrm{a}}} \mathcal{L_C}\left(\hat{y}_{i, s}^{c}, y_{i}\right)
 \end{equation}
  \begin{equation}
     \mathcal{L}_{p a}=-\frac{1}{|{D^{train}_t}|} \sum_{x_{i} \in {D^{train}_t}} \log \frac{1}{\mathrm{T}_{\mathrm{a}}} \sum_{s=1}^{\mathrm{T}_{\mathrm{a}}} \mathcal{L_D}\left(\hat{y}_{i, s}^{p}, p_{c}\right)
 \end{equation}

 where $\sigma_i^c$ and  $\sigma_i^p$ are the standard deviation obtained from Eq.~\ref{eq:var_c} and Eq.~\ref{eq:var_p}. $T_a$ is the number of Monte Carlo (MC) samples. The models are trained by jointly minimizing both the classification loss and aleatoric loss. The total aleatoric loss is given as:
 \vspace{-0.5em}
 \begin{equation}
 \mathcal{L}_{ale}= \mathcal{L}_{c a} +\mathcal{L}_{p a}
 \label{eq:ale}
 \end{equation}
 \vspace{-3em}
\subsubsection{Distillation using aleatoric uncertainty}
The objective of incremental learning is to preserve the previous task information while training for the current tasks. Thus preserving uncertainty information provided by the  Bayesian neural network on the previous task can also overcome the limitation of prediction-based distillation loss. In this work, we propose an aleatoric uncertainty distillation loss to transfer the previous task's information to the current model. The aleatoric uncertainty tells us how much the model is uncertain about the prediction due to the data. In the incremental learning setting, the previously trained model's data uncertainty should be similar for the current task models. This is because the data uncertainty indicates properties such as low illumination or blur that would be a factor of uncertainty for all methods. In incremental learning, at task $t$, the current model $M_t$ is initialized with the previous task model $M_{t-1}$. The objective is to train the $M_t$ model from the current task dataset i.e. $D^{train}_t$. We use the cross-entropy loss to train $M_t$ from the current task data.
For any input image, $x$ of current task data $x \in D^{train}_t$ is forwarded to the current model and previous model to predict the aleatoric variance value.  For transferring the uncertainty, we define the aleatoric distillation loss $\mathcal{L_A}$  using the following equation:

\begin{equation}
   \mathcal{L_A}= \sum_{i} || (\sigma_{i, {t-1}}^{p})^2-  (\sigma_{i, {t}}^{p})^2||^2
    \label{eq:ale_mse}
\end{equation}

\subsection{ Attention based Distillation}
The basic intuition is for attention-based distillation is that not all information needs to be preserved for the next task. In the traditional incremental learning, the method that has proposed the distillation loss apply it to preserve the full image information. It reduces a model's capability for learning a new task as it is learning unnecessary information. In ~\cite{dhar2019learning}, the authors use a GradCAM activation to distill the model information. These activations are obtained by the network's prediction, which is overconfident for even for the wrong prediction due to mismatch of train-test distribution~\cite{hendrycks2019augmix}, and thus do not capture the full prediction distribution.
In contrast to the above global and activation based distillation, we are training the model to learn the attention map itself. Inspired by the self-attention methods~\cite{vaswani_nips2017,zhang2019self}, we introduce an attention network to predict the attention map. These attention models are trained by  the residual information to classification loss. Let the attention model $Att(:,\theta_{att})$ be parametrized by  $\theta_{att}$, then the attention obtained similar to~\cite{zhang2019self} is given by:
\begin{equation}
     y^{att}_{i,t}=Att(x^{f}_{i,t}, \theta_{att,t})
\end{equation}
  $(x^{f}_{i,t})$ is the feature representation at task $t$ for input $x_i$.
\subsubsection{Attention Distillation loss}
In the incremental learning setting, if the current model $M_{t}$ and previous model $M_{t-1}$  have equivalent knowledge of base classes, they should have a similar response to these regions. Therefore the attention map obtained by the previous model and the current model should be similar. It transfers the old class data information to the current model. Since the previous model is trained on the old task data, the similarity between the previous model and the current model's attention map for new task data provides us with a signature of the old task data. So the current model objective is to preserve these signatures while learning the new task.  We use the mean square loss for the attention distillation. $L_2$ loss or mean squared loss is obtained by squaring the difference between attention from the previous model and the current model. The loss function is defined as follows:
\begin{equation}
   L^{t}_{att}(y^{att}_{i,t-1},y^{att}_{i,t})= \sum_{i} ||  y^{att}_{i,t-1}-  y^{att}_{i,t}||^2
    \label{eq:mse}
\end{equation}

\subsubsection{Multi Stage Attention Distillation}
In the convolution neural network, each stage learns a different feature for an input image. We use the attention module in each stage to predict the attention or essential part of images. These stage-wise features reveal the semantic structure of these images. In the proposed framework, we use these stage activations of the network to generate the attention map.  For the incremental learning setting, we distill this attention map from the older model to the current model to better learns to preserve the older task information. In the proposed framework, we consider 3 stage self-attention net in the incremental class setting. We provide the visualization of the 3rd stage attention map in  Fig.~\ref{fig:att_vis}.
The multi-stage attention distillation loss is given as: 
\begin{equation}
    \mathcal{L}_m=\sum_{s=1}^3 L^{t}_{att}(y^{att(s)}_{i,t-1},y^{att(s)}_{i,t})
    \label{eq:multi}
\end{equation}
Here $y^{att(s)}_{i,t-1}$ corresponds to the attention map of $i^{th}$ input at step $t$ from the $(t-1)th$ model at stage $s$.

\subsection{Total Loss}
We train the model using the cross entropy loss (Eq.~\ref{eq:crs}),  distillation loss (Eq.~\ref{eq:dis}), aleatoric uncertainty (Eq.~\ref{eq:ale}), uncertainty distillation (Eq.~\ref{eq:ale_mse}) and  attention distillation loss (Eq.~\ref{eq:multi}). The total loss is given as follows:

\begin{equation}
    L^{t}_{total}=  \lambda (\mathcal{L}_m + \mathcal{L_A}+ \mathcal{L}_{ale}) + \mathcal{L}_{\mathrm{C}}+  \mathcal{L}_{\mathrm{D}}
\end{equation}
Here $\lambda$ is the weighting parameter for the uncertainty and attention based distillation.

 \begin{table}[t]
            \centering
            \scalebox{0.9}{
                \begin{tabular}{|l|c|c|c|c|c|c|} \hline
                \multirow{2}{*}{\textbf{Method}} & \multicolumn{2}{c|}{5 class} & \multicolumn{2}{c|}{10 class} & \multicolumn{2}{c|}{20 class} \\ \cline{2-7} 
                &    ACC   &    FGT    &      ACC  &   FGT&     ACC  &       FGT  \\ \hline
                \textbf{Baseline}                    &    57.4&21.0 &56.8&19.7&56.0& 18.0 \\
                LwF-MC\cite{li2017learning} &   58.4&19.3 &59.5&16.9&60.0&14.5\\
                D+R\cite{hou2018lifelong}&   59.1& 19.6 &60.8&17.1&61.8&14.3\\
                E2E\cite{castro2018end} &     60.2&16.5&62.6& 12.8&65.3& 8.9\\
                
                 iCaRL\cite{rebuffi2017icarl}* &     61.2&-&64.1& -&67.2& -\\
                
                 GD\cite{lee2019overcoming} &     {62.1}&15.4&65.0& 12.1&67.1& 8.5\\
                 MD~\cite{zhao2020maintaining}  &  \textbf{62.6}&- & 64.4&- & 66.6 & - \\

                \hline
                Ours-AU& {61.8} & \textbf{15.2}   & \textbf{65.4}& \textbf{11.4}& \textbf{67.6}&\textbf{8.2} \\ 
                \hline
                \textbf{Oracle} &    78.6&3.3&77.6&3.1&75.7&2.8\\
                \hline
                \end{tabular}
                }
                \caption{Results on CIFAR-100. Baseline refers when distillation loss in not used and  Oracle refers when we store full data samples. iCaRL uses ResNet-32 backbone.}
                \label{tbl:cifar100}
    \end{table}

\section{Experiments and Analysis}
We evaluate our proposed method by comparing it with state-of-the-art methods for dataset CIFAR-100  and Imagenet dataset. The ablation analysis with uncertainty and attention-based distillation is also shown in subsequent sections. Also, we have provided attention visualization for various methods, as shown in  Fig.~\ref{fig:att_vis}. For our evaluation purpose, we use CIFAR-100 and Imagenet dataset, as mentioned in section~\ref{dataset}.

\subsection{Dataset}
\label{dataset}
\textbf{{CIFAR-100}:} In the CIFAR-100~\cite{krizhevsky2009learning} dataset, there are 100 classes containing 60,000 RGB images of size 32 x 32. Each class contains 500 training images and 100 test images. We follow the benchmark protocol given in~\cite{rebuffi2017icarl} for our experiments. The classes are shuffled similar to~\cite{rebuffi2017icarl,castro2018end}. We train all 100 classes of CIFAR-100 in incremental batches of 5, 10, and 20 classes, so respectively, there are 20,10 and 5 tasks present. 

\noindent\textbf{ImageNet:} For the Imagenet dataset~\cite{deng2009imagenet}, we follow a similar approach provide in~\cite{lee2019overcoming} by sample 500 images per 100 randomly chosen classes for each trial, and then split the classes.  These images also have a dimension of 32 x 32. We use the same split of~\cite{lee2019overcoming} in our paper. In ImageNet, we also used incremental batches of 5, 10, and 20 classes, so respectively, there are 20, 10, and 5 tasks present.

\subsection{Implementation Details}
We follow similar protocols, as discussed in \cite{lee2017overcoming}.
We use Wide-ResNet~\cite{zagoruyko8wide} architecture as our base model.  We also obtain a comparable performance with ResNet~\cite{he2016deep}. The model is implemented in the PyTorch~\cite{paszke2017automatic} framework. The aleatoric variance layer is a linear layer followed by the soft plus layer. The attention modules are consist of convolution layers. All the models are initialized with random weights.  During training, we use a batch size of 128 in all the experiments. We standardize input images by subtracting them with the mean of the training set and dividing them by the standard deviation. In all methods,  we use the total buffer size of the representative image number of 2000~\cite{rebuffi2017icarl,lee2019overcoming}.  These representative images are randomly selected. Other details are provided in the project page\footnote{\url{https://delta-lab-iitk.github.io/Incremental-learning-AU/}}.

 \begin{figure*}[!]
     \small
     \centering
     \begin{tabular}[b]{ c c   }
     (a)Accuracy plot for 5 class  increment & (b)Forgetting for 5 class increment \\ 
     \includegraphics[width=0.43\textwidth]{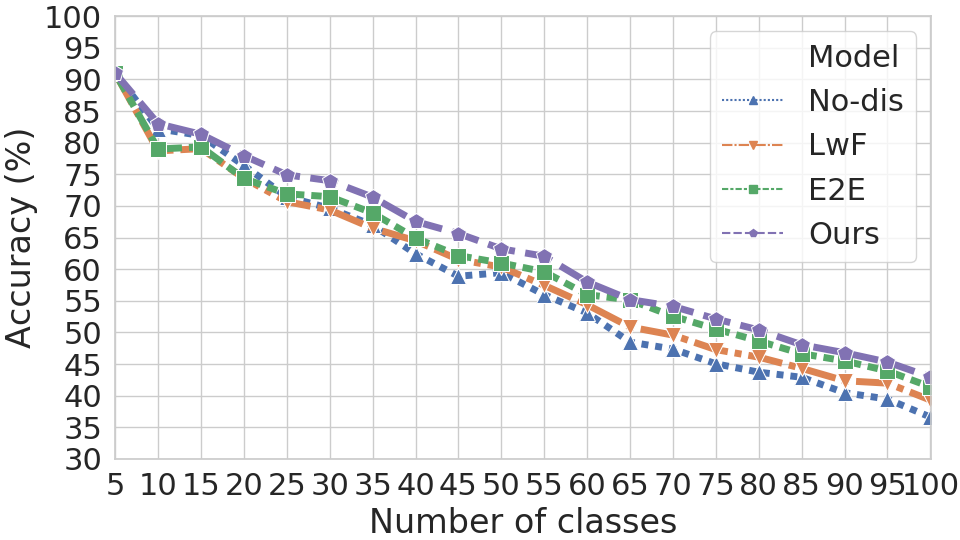}
     & \includegraphics[width=0.43\textwidth]{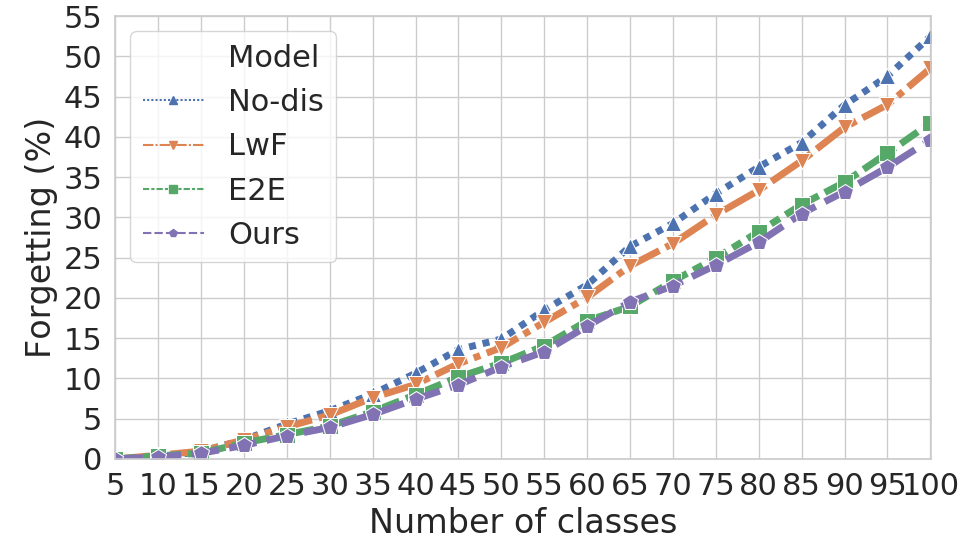} \\
     (c)Accuracy plot for 10 class  & (d)Forgetting for 10 class increment \\ 
     \includegraphics[width=0.43\textwidth]{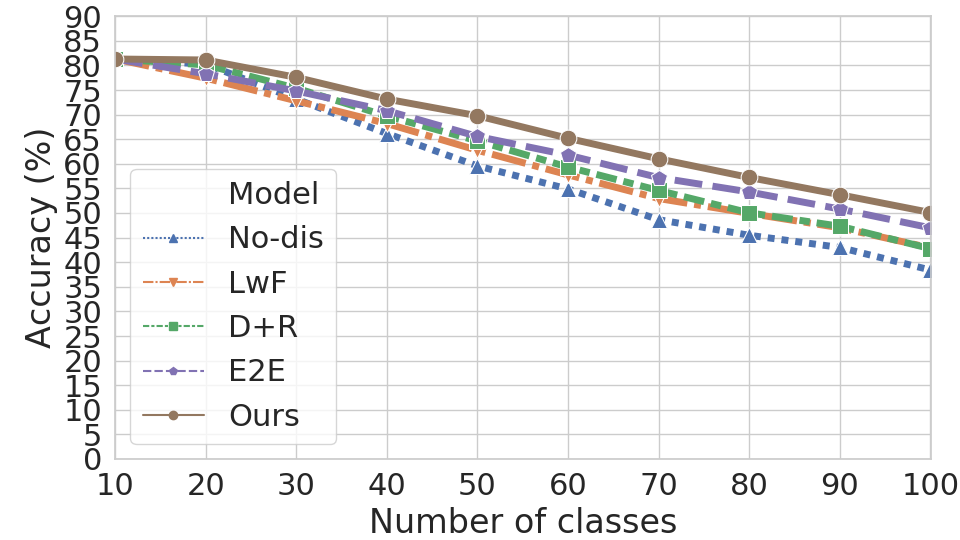}
     & \includegraphics[width=0.43\textwidth]{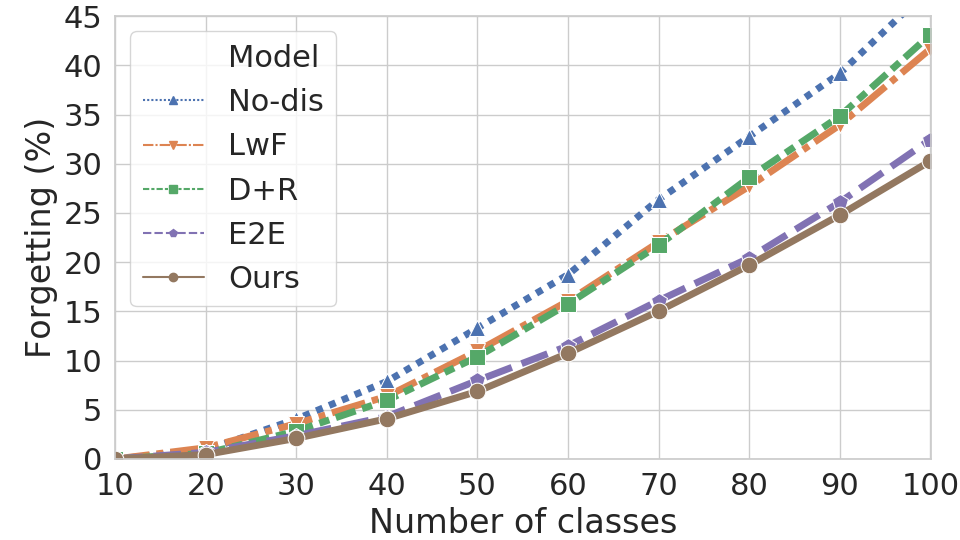} \\
     (e)Accuracy plot for20 class & (f)Forgetting for 20 class increment \\ 
      \includegraphics[width=0.43\textwidth]{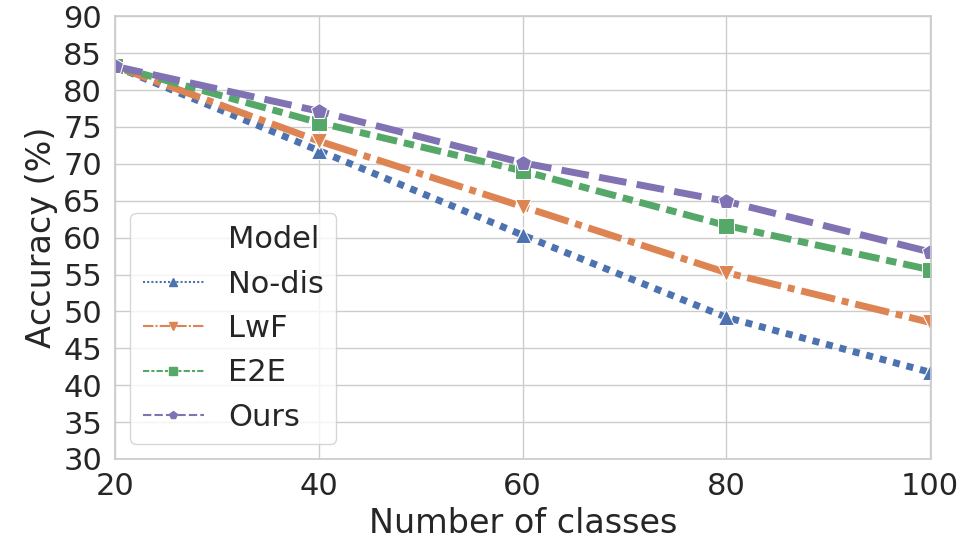} 
       &\includegraphics[width=0.43\textwidth]{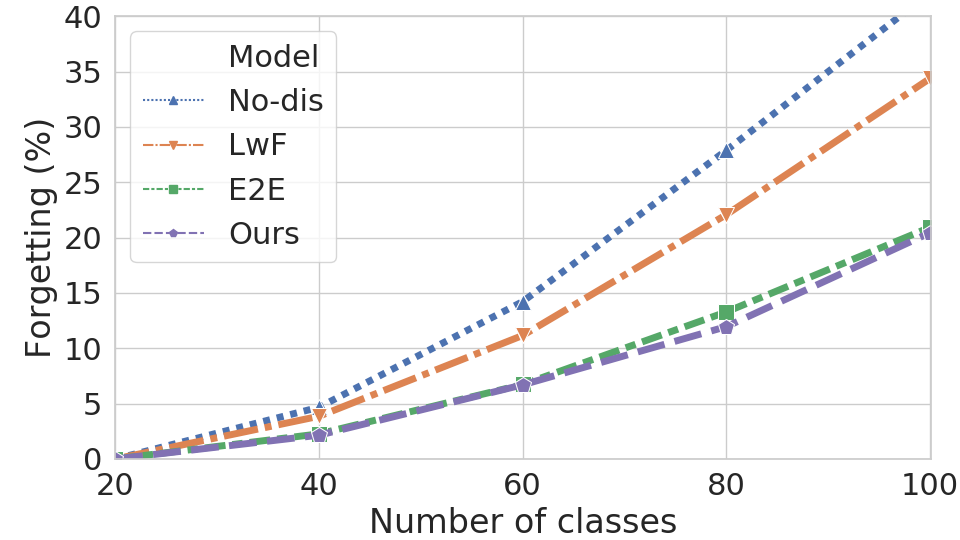} \\
   \end{tabular}
      \caption{This figure shows accuracy and forgetting plot for 5 class , 10 class and 20 class increments for CIFAR-100 dataset }
      \label{fig:class_incre}
 \end{figure*}

\subsection{Comparison with state-of-the-art methods}
\label{SOTA}
We provide comparison with other state-of-the-art methods such as Learning without forgetting (LwF)~\cite{li2017learning}, distillation and retrospection (D+R)~\cite{hou2018lifelong}, End-to-End incremental learning (E2E)~\cite{castro2018end},  Global Distillation (GD)~\cite{lee2019overcoming}, iCaRL~\cite{rebuffi2017icarl} and  maintaining  discrimination (MD)~\cite{zhao2020maintaining}. The upper bound results correspond when we consider all the task data. Baseline results, when we are not using any distillation and uncertainty estimation.  The CIFAR-100 results are provided in Table~\ref{tbl:cifar100} with increment tasks size 5, 10, and 20. The state-of-the-art method's values in the table have been evaluated from~\cite{lee2019overcoming}. In  Fig.~\ref{fig:class_incre}, we have plotted the accuracy and forgetting of these increment tasks. Similarly, ImageNet dataset's results are also provided in Table~\ref{tbl:imagenet} with increment tasks size are 5, 10, and 20. 
In both of the tables, the baseline refers to the model where distillation loss is not used;  oracle refers to when we store full data samples. ACC refers to average accuracy (higher is better), and FGT refers to the amount of catastrophic forgetting, by averaging the performance decay (lower is better). We followed a similar approach of~\cite{lee2019overcoming} to obtain the ACC and FGT.
Our Attention Uncertainty (AU) methods improves~4\% from the baseline method~\cite{li2017learning} in 5 class increment steps, around 6\% improvement as in 10 class increment steps, and around 7\% improvement in terms of accuracy in 20 class increment steps. Also, we have ~1.6\% improvement in accuracy in comparison with the state of the art method~\cite{castro2018end} in 5 steps, ~2.8 \% in 10 steps, and 1.5\% in 20 steps. Note that this additional improvement in accuracy is accompanied by the ability to interpret the model through the obtained by uncertainty estimates and the attention map. 

 \begin{table}[h!]
        \centering
                \begin{tabular}{|l|c|c|c|c|c|c|} \hline
                \multirow{2}{*}{\textbf{Method}} & \multicolumn{2}{c|}{5 class} & \multicolumn{2}{c|}{10 class} & \multicolumn{2}{c|}{20 class} \\ \cline{2-7}  
                &    ACC   &    FGT    &      ACC  &   FGT&     ACC  &       FGT  \\ \hline
                \textbf{Baseline}   &    44.2&23.6 &44.1&21.5&44.7& 18.4 \\
                LwF-MC &   45.6&21.5 &47.3&18.5&48.6&15.3\\
                D+R&   46.5& 22.0 &48.7&18.8&50.7&15.1\\
                E2E &     47.7&17.9&50.8& 13.4&53.9& 8.8\\
                
                GD &     50.0&16.8&53.7& 12.8&56.5& 8.4\\
                \hline
                Ours-AU &  \textbf{51.2}  & \textbf{15.4}&  \textbf{54.3}& \textbf{11.9} &\textbf{56.9} & \textbf{ 8.1}\\ 
                \hline
                Oracle &    68.0&3.3&66.9&3.1&65.1&2.7\\
                \hline
                \end{tabular}
                \caption{Results on ImageNet-100. Baseline refers when distillation loss in not used and  Oracle refers when we store full data samples}
                \label{tbl:imagenet}
                \vspace{-2em}
          \end{table}

\subsection{Class Incremental Accuracy}
\label{class}
For showing the class increment effect in incremental class learning, we have plotted each incremental step average accuracy as well as forgetting in  Fig.~\ref{fig:class_incre} for CIFAR-100 dataset. The  Fig.~\ref{fig:class_incre}(a) shows for 5 class increments, (b) for 10 class increments, and (c) for the 20 class increments.  Similarly, for the same CIFAR-100 dataset, we have plotted the forgetting value for the different models on (d) 5 class increments, (e) for 10 class increments, and (f) for the 20 class increment task in the  Fig.~\ref{fig:class_incre}.
From the figure, it is clear the proposed uncertainty, and attention distillation based method outperforms as compare to other state-of-the-art methods Learning without forgetting (LwF)~\cite{li2017learning}, Progressive distillation and retrospection (D+R)~\cite{hou2018lifelong}, iCaRL~\cite{rebuffi2017icarl}, End-to-End incremental learning (E2E)~\cite{castro2018end} and Global distillation (GD)~\cite{lee2017overcoming}.

 \begin{figure}
 \centering
    \includegraphics[height=10cm,width=0.420\textwidth]{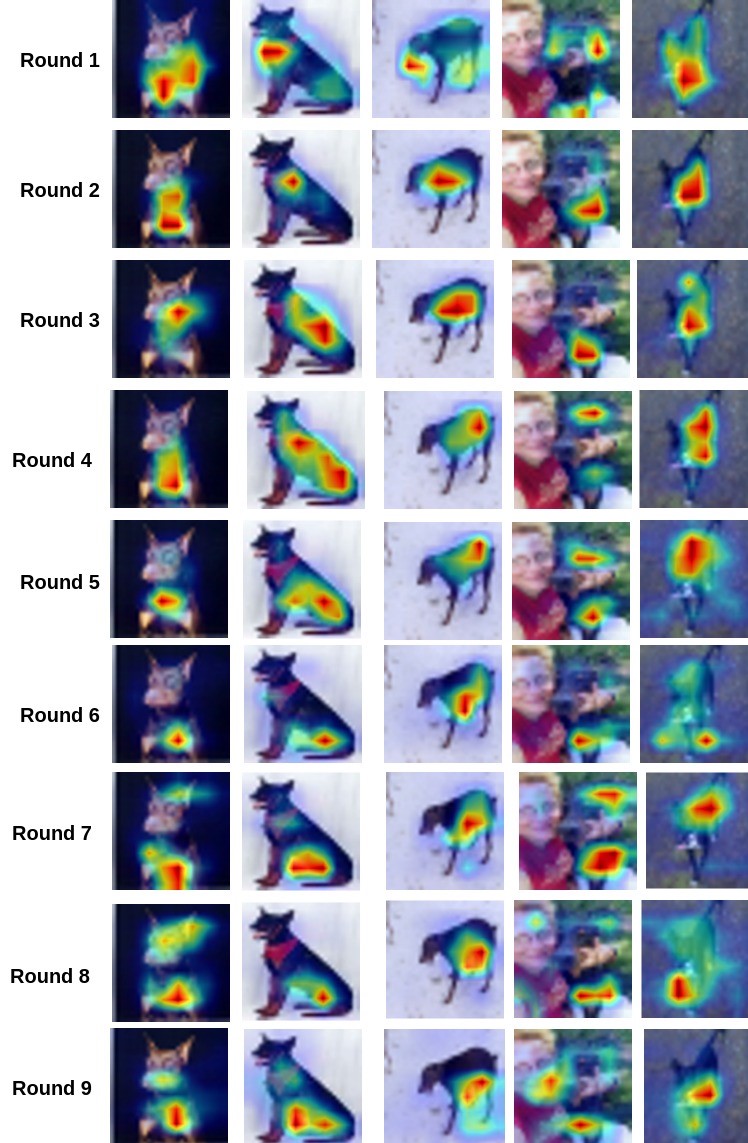}
      \caption{Visualization of attention map in CIFAR-100 in different task learning stage of 10 class increment setting. Each row represents the task stage, }
      \label{fig:att_vis}
 \end{figure}

\begin{table}[!h]
            \centering
    \scalebox{1}{
       \begin{tabular}{|c|c|c|}
\hline
 Methods& ACC & FGT  \\ \hline
 AU &  63.81& 12.22 \\
 AD & 64.37 &  12.03 \\ 
 AU +UD & 64.54 &  11.84 \\ 
 AU + AD & 64.52 &  11.92 \\ 
 AU +UD + AD& \textbf{65.42} & \textbf{11.43}  \\  \hline
\end{tabular}
}
\caption{Ablation analysis for different loss functions on CIFAR-100 dataset for 10 class incremental setup.}
\vspace{-1em}
\label{tbl:abl_cifar}
    \end{table}

 \begin{table}[!h]
        \centering
  \scalebox{0.9}{
       \begin{tabular}{|c|c|c|c|}
\hline
               $\lambda$  &Model  & ACC & FGT \\ \hline
\multirow{2}{*}{0.1} & AU + AD & 64.32 &11.97  \\ \cline{2-4} 
                  & AU +UD + AD &64.63  & 11.76 \\ \hline
\multirow{2}{*}{0.5} & AU + AD & 64.54 &  11.84 \\ \cline{2-4} 
                  & AU +UD + AD  & \textbf{65.42} &  \textbf{11.43}\\ \hline
\multirow{2}{*}{1} & AU + AD & 64.13 & 12.11 \\ \cline{2-4} 
                  & AU +UD + AD  &  64.25& 12.08 \\ \hline
\end{tabular}
}
\caption{Ablation analysis for different weights of loss functions on CIFAR-100 dataset for 10 class incremental setup. }
\label{tbl:cifar_weight}
    \end{table}
 
\subsection{Effect of Attention and Uncertainty Distillation }
To analyze the effect of uncertainty and attention in incremental learning, we provide the ablation analysis of different loss functions in Table~\ref{tbl:abl_cifar} for the CIFAR-100 dataset for 10 class increment setup. Here, for all, we use the $\lambda=0.5 $  for uncertainty and attention distillation loss. In Table, \textbf{AU:} Aleatoric Uncertainty,\textbf{ AD:} Attention Distillation and \textbf{UD:} Uncertainty Distillation. In this experiment, we can see that the uncertainty distillation and attention distillation improve the model performance. 

\subsection{Effect of weights of Attention and Uncertainty Distillation loss}
To further analyze the effect of attention and uncertainty on the incremental learning setup, we experiment with different weight values for the loss, and the results are reported in Table~\ref{tbl:cifar_weight}. In the table, \textbf{$\lambda$} refers to the weight value of uncertainty and attention distillation.  This experiment is performed for the CIFAR-100 dataset and 10 incremental class setup. We experiment with weight value for 0.1, 0.5, and 1. We can see that when the distillation loss weighted by 0.5 gives better performance in terms of better accuracy and less forgetting. This is expected because as we increment the distillation loss's contribution, it does not perform well on the current task; similarly, if the weight value is less, it does not keep the previous task information very well. 

\subsection{Attention Visualization}
\label{vis}
In  Fig.~\ref{fig:att_vis}, we provide the attention map visualization for a given input of different class CIFAR-100 dataset in 10 class increment setting.  In each column, we have shown different instances of an example, while along the column, we have shown the attention visualization in different step sizes. For example, in the first row, the attention map is generated when the model generated an attention map for a model trained on the first 10 classes. Similarly, the second row shows the attention after the model is trained on till 20 classes in increment setting. Note that visualization is shown here from the last stage of the attention module. From the figure, we observe that the attention map is located in the object region for each task. The model distills the attention information to the next task; this helps to avoid the catastrophic forgetting in incremental learning.

\section{Conclusion}

In this paper, we consider the task of incremental learning, where we aim to have models that are able to preserve their performance across tasks. Several methods have been proposed for this task; these have mostly considered point estimates during prediction; for instance, those obtained using softmax. We show that these methods are susceptible to be incorrectly certain about performance, and by incorporating uncertainty, we are able to obtain a method that incrementally learns on tasks while also accurately providing the uncertainty of the model. Further, by incorporating self-attention, we can also visualize the regions on which the method attends across tasks. Further, our results show consistent improvements over the state of the art methods on standard datasets. 

\clearpage

{\small
\bibliographystyle{ieee_fullname}
\bibliography{egbib}
}

\end{document}